\documentclass[letterpaper]{article}
\usepackage{aaai2026}
\usepackage{times}
\usepackage{helvet}
\usepackage{courier}
\usepackage[hyphens]{url}
\usepackage{graphicx}
\usepackage{natbib}
\usepackage{caption}
\usepackage{algorithm}
\usepackage{algorithmic}
\usepackage{subcaption}
\usepackage{graphicx}
\usepackage{enumitem}
\usepackage{xcolor}
\usepackage[most]{tcolorbox}

\tcbset{
  grayboxstyle/.style={
    colback=gray!10,
    colframe=gray!80!black,
    boxrule=0.6mm,
    arc=3mm,
    width=\linewidth,
    enhanced,
    breakable
  }
}

\usepackage{booktabs}
\usepackage{multirow}
\usepackage{amsmath}
\usepackage{amssymb}

\newcommand{\ourmethod}{EffiEval}

\usepackage{newfloat}
\usepackage{listings}
\DeclareCaptionStyle{ruled}{labelfont=normalfont,labelsep=colon,strut=off} 
\floatstyle{ruled}
\newfloat{listing}{tb}{lst}{}
\floatname{listing}{Listing}

\title{\ourmethod{}: Efficient and Generalizable Model Evaluation\\ via Capability Coverage Maximization}
\author{
    Yaoning Wang\textsuperscript{\rm 1},
    Jiahao Ying\textsuperscript{\rm 2},
    Yixin Cao\textsuperscript{\rm 1}\thanks{Corresponding Author},
    Yubo Ma\textsuperscript{\rm 3},
    Yugang Jiang\textsuperscript{\rm 1}
}
\affiliations{
    \textsuperscript{\rm 1}Institute of Trustworthy Embodied AI, Fudan University\\
    \textsuperscript{\rm 2} Singapore Management University\\
    \textsuperscript{\rm 3} Nanyang Technological University\\
    24210240050@m.fudan.edu.cn,  yxcao@fudan.edu.cn

}
\begin{document}

\nocopyright
\maketitle

\begin{abstract}
The rapid advancement of large language models (LLMs) and the development of increasingly large and diverse evaluation benchmarks have introduced substantial computational challenges for model assessment. In this paper, we present \ourmethod{}, a training-free approach for efficient benchmarking that effectively addresses data redundancy while maintaining high evaluation reliability. Our method is specifically designed to meet three key criteria for high-quality evaluation: \textbf{representativeness}, by ensuring comprehensive coverage of model capabilities; \textbf{fairness}, by remaining independent of model performance during sample selection to avoid bias; and \textbf{generalizability}, by enabling flexible transfer across datasets and model families without reliance on large-scale evaluation data. Unlike traditional methods that rely on absolute performance or require extensive evaluation data, our approach adaptively selects high-quality representative subsets based on the Model Utility Index (MUI). Extensive experiments on multiple public benchmarks and diverse LLMs demonstrate that \ourmethod{} achieves strong ranking consistency with full-dataset evaluation using only a small fraction of the original data. Furthermore, our method is flexible and scalable in size, allowing users to balance evaluation efficiency and representativeness according to specific needs. Overall, \ourmethod{} provides a practical and generalizable solution for reliable, fair, and efficient evaluation in the era of LLMs.\footnote{\url{https://castria-cn.github.io/EffiEval-homepage/}}
\end{abstract}

\section{Introduction}
Based on the training scaling laws~\cite{kaplan2020scaling}, large language models (LLMs) are becoming significantly more capable as computational resources, model parameters, and training data scale up continuously. This trend has driven researchers to construct increasingly large and diverse benchmarks for comprehensive model evaluation, such as MMLU~\cite{hendrycks2020measuring}, HELM~\cite{liang2022holistic}, and BIG-Bench~\cite{srivastava2023beyond}. However, the scale of these benchmarks introduces substantial evaluation costs. For example, on the HELM benchmark, evaluating a single model can require over 500 GPU hours~\cite{liang2022holistic}.
This computational burden is further increased by the growing adoption of test-time scaling~\cite{openai2024gpt4o, guo2025deepseek}, where longer inference times are used to boost performance.
Worse still, the rapid iteration and frequent updating of LLMs further intensify this evaluation cost.
Therefore, improving evaluation efficiency while ensuring high quality has become an increasingly important challenge in the era of LLMs.

We assume that traditional evaluation datasets contain a certain degree of redundancy. Therefore, it is possible to down-sample the data to achieve Efficient Benchmarking~\cite{perlitz2023efficient} --- that is, to intelligently reduce the computational cost of evaluation without compromising its reliability. This reliability can be defined in two ways: 1) the absolute value of performance metrics remains the same, which is emphasized in previous works~\cite{polo2024tinybenchmarks, kipnis2024metabench}, or
2) the ranking of multiple models is preserved. We argue that preserving the absolute values of performance metrics is unnecessary. First, absolute scores are more sensitive to data distribution shifts. Suppose a model performs particularly well on coding questions, which make up 70\% of the dataset. If the proportion of coding questions decreases, the model's absolute performance score will naturally decline. However, such changes do not necessarily indicate that the selected subset is of poor quality --- it may simply reflect a shift in the data distribution. Second, scores are affected by the difficulty of the sample. When easy questions are removed and the difficulty gap is widened, absolute scores will inevitably drop. Therefore, we adopt the relative ranking among multiple models as the primary measure of evaluation consistency.

At the same time, a high-quality subset should also meet additional criteria to ensure that it remains representative, fair, and generalizable. Specifically:
1) It should still cover the diverse capabilities of models as much as possible to ensure comprehensive evaluation. For example,~\cite{perlitz2023efficient} apply stratified random sampling based on the scenarios defined in the original benchmark. However, this approach is heavily influenced by the original data distribution, making it difficult to adequately sample from sparse categories or domains, which may even be entirely missed during sampling.
2) It should be uncorrelated with model performance, to avoid introducing bias. Clearly, when the sampling process is correlated with model performance --- such as selecting samples where models differ the most --- it may introduce evaluation bias. 
3) It should exhibit a certain level of generalizability. For example, if generating the subset requires evaluating all data on all evaluated models beforehand and cannot adapt to new model evaluations, then it does not truly improve evaluation efficiency. Previous statistic-based approaches~\cite{polo2024tinybenchmarks, kipnis2024metabench} rely on large amounts of evaluation data to select informative samples, making them difficult to transfer to new datasets or adapt to unseen evaluation settings.

In this paper, we propose a training-free efficient benchmarking method, \textbf{\ourmethod{}}, which satisfies the above three criteria. Inspired by the Model Utility Index (MUI)~\cite{cao2025model}, which introduces an evaluation metric to quantify a model’s ``effort'' on a given dataset, our core idea is to reduce data redundancy by maximizing the number of activated neurons from the model’s internal mechanism, while simultaneously preserving the diversity of covered capabilities.
Unlike traditional diversity-based criteria (e.g., domains or predefined capabilities), our method is model-specific --- it selects evaluation samples that are diverse with respect to a given model, ensuring a broad and representative assessment. 
To mitigate the performance bias issue, our approach does not rely on large-scale evaluation data to train sample representations or performance predictors. This ensures that the selection process remains independent of the model’s actual performance, while also being efficient and easily transferable to new datasets.
In later experiments, we demonstrate the generalizability of our selected subsets: subsets chosen based on one model can also provide reliable evaluation results for other models. We argue that, despite differences in training data and architectures, many models share similar distributions of capability diversity, which contributes to the observed generalization. In this sense, our approach can also be viewed as a form of meta-evaluation, enabling a quantitative assessment of dataset redundancy or diversity.

In our experiments, we observe the following: 1) Using as little as 5\% of the original data, our method achieves an average Kendall’s $\tau$ greater than 0.9 across multiple benchmarks, indicating strong preservation of evaluation rankings; 2) When increasing the subset to 10\%, the average Kendall’s $\tau$ exceeds 0.95, reflecting even stronger consistency with full-data evaluation; 3) Unlike prior approaches that require predefining the subset size or searching across all possible sizes, our method dynamically determines the subset size based on the desired coverage or performance correlation. 

Our contributions can be summarized as follows:

\begin{itemize}
    \item We highlight the task of efficient benchmarking and argue that a high-quality subset must meet several key requirements to ensure representativeness, fairness, and generalizability.
    \item We propose a training-free subset sampling method \ourmethod{} that balances evaluation efficiency and data representativeness, without relying on large-scale evaluation data, thus enabling easy transferability to other datasets.
    \item Extensive experiments demonstrate that our method can adaptively select a representative subset that not only covers diverse model capabilities but also preserves the performance ranking among models.
\end{itemize}
\section{Related Work}
\begin{description}[style=unboxed, leftmargin=0cm]
  \item[\textbf{Efficient and Generalizable Evaluation}.] As LLM capabilities grow rapidly, a key challenge is that fixed test datasets and static metrics can no longer keep up~\cite{cao2025generalizableevaluationllmera}. New models may outperform existing benchmarks without revealing their full potential, or appear weaker simply because the tests fail to capture their emerging abilities. This growing mismatch calls for more generalizable and adaptive evaluation methods. To address this, prior work has explored various approaches for evaluating LLMs efficiently and reliably. One common direction takes the evaluator perspective, where LLMs themselves are leveraged as evaluators to reduce the cost of data construction and annotation, and to enable more up-to-date and scalable evaluation~\cite{NEURIPS2023_f64e55d0, NEURIPS2024_1e89c126}. Another line of work focuses on estimating model performance based on low-cost proxies, such as model size and training token count (i.e., scaling laws~\cite{kaplan2020scaling}), or performance on a carefully selected, representative subset of data~\cite{polo2024tinybenchmarks, pacchiardi2024100, kipnis2024metabench}—the latter being a form of evaluation data selection.
  \item[\textbf{Evaluation Data Selection}.] Evaluating models on large benchmarks is time- and resource-intensive. To address this, several studies have proposed selecting a representative subset for evaluation and extrapolating the full dataset performance. For example,~\cite{vivek2023anchor} leverage confidence scores on classification benchmarks to select samples with the highest correlation in scores with the rest of the dataset. ~\cite{polo2024tinybenchmarks} fit an Item Response Theory (IRT) model using prior LLM performance, then apply K-Means clustering on the estimated item parameters to select representative samples.~\cite{pacchiardi2024100} follow a similar pipeline but use sample embeddings obtained from the OpenAI API. ~\cite{kipnis2024metabench} also adopts an IRT-based approach but uses Fisher Information to filter out less discriminative samples. These methods rely on extensive performance data (from 400 to 5000 models on a single dataset), causing substantial computational overhead and potential bias. In contrast, our method requires minimal evaluation data yet still generalizes well, achieving high correlation between subset and full-dataset performance.
\end{description}
\section{MUI Based Evaluation Data Selection}

\begin{figure*}[htbp]
  \centering
  \includegraphics[width=0.95\textwidth,page=1]{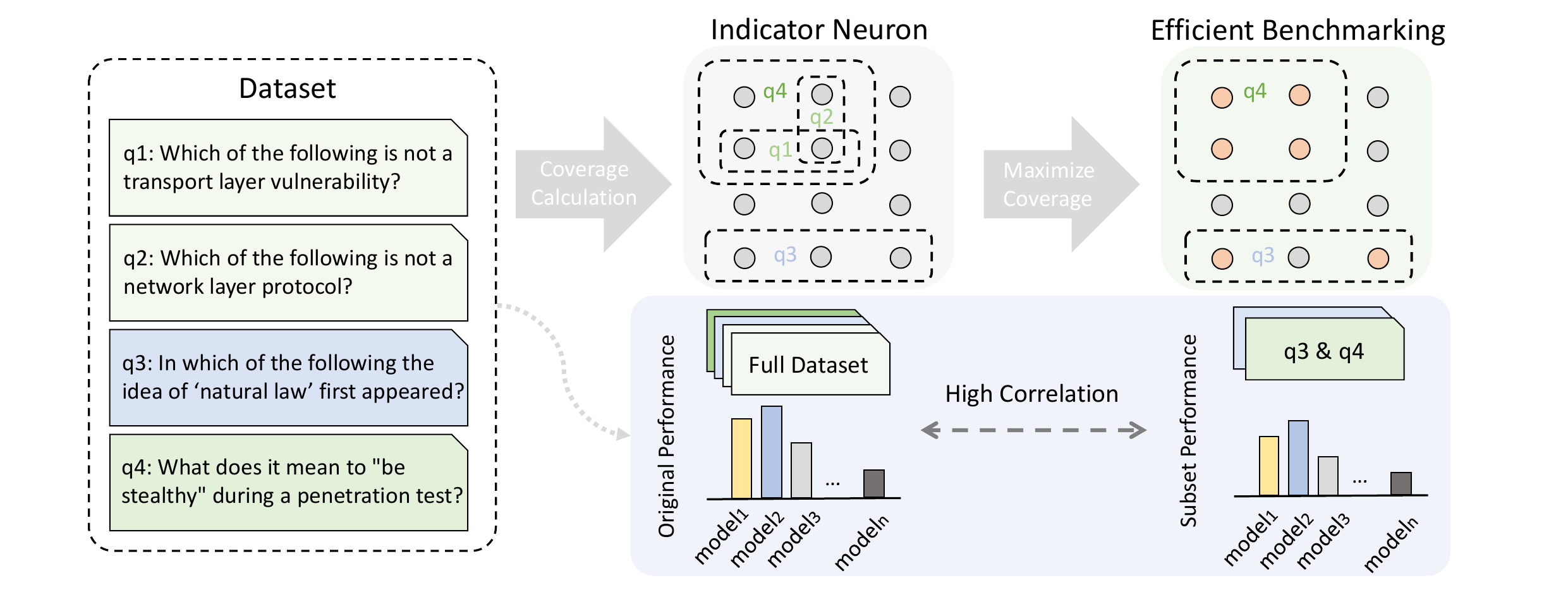}
  \caption{Framework of our method \textbf{\ourmethod{}}. By computing the evaluation coverage of the indicator model, we select samples that maximize the set of covered capabilities, thereby constructing a representative evaluation subset as a substitute for the full benchmark. To maximize the coverage of model capabilities, we select a subset that activates the broadest range of indicator neurons. In this example, $\{q_3, q_4\}$ is selected due to its diverse neuron activation patterns. The detailed algorithm is in Algorithm~\ref{alg:algorithm}.}
  \label{fig:workflow}
\end{figure*}

\subsection{MUI Computation}
Inspired by prior efforts to quantify model efficiency, the Model Utility Index (MUI)~\cite{cao2025model} is proposed as the foundation of our work to measure the amount of effort a model expends to achieve a given outcome. The MUI is defined as:
\begin{equation}
\label{eq:MUI}
\textbf{MUI}(t)=\frac{N_\text{activated}(t)}{N_\text{total}}
\end{equation}
where $N_\text{total}$ denotes the total number of capabilities (e.g., neurons or features) of the model, and $N_\text{activated}(t)$ represents the number of activated capabilities when the model completes task $t$.
Built upon interpretation techniques, MUI can naturally measure the extent to which a model’s capabilities are exercised by specific tasks, as studies~\cite{pan-etal-2024-finding, templeton2024scaling} have shown that different types of knowledge and abilities are associated with distinct sets of key neurons or features. To ensure broader applicability, in this work, we adopt a neuron-based MUI calculation for all experiments. Specifically,
Given an input sample $x$ and its corresponding model prediction $y = (\hat{y}_1, \hat{y}_2, \ldots, \hat{y}_t)$, \citet{cao2025model} defines the neuron-based contribution score $f_{\text{neuron}}(i,l,\hat{y}_j \mid x) \in \mathbb{R}$ of the $i$-th neuron in layer $l$ to the prediction of token $\hat{y}_j$ as
\begin{equation}
\begin{aligned}
f_{\text{neuron}}(i,l,\hat{y}_j \mid x) = \left(\mathbf{W}_u\mathbf{W}_{\text{out}}^{l}\circ\sigma\left(\mathbf{W}_{\text{in}}^{l} \left(\mathbf{x}^l_{-1}\right)\right)^{\top}\right)_{i,\hat{y}_j},
\end{aligned}
\label{eq:neuron_contribution}
\end{equation}
where $\sigma$ is an activation function, $\mathbf{W}_{\text{in}}^{l}$ and $\mathbf{W}_{\text{out}}^{l}$ are the input/output projections in FFN, $\mathbf{W}_u$ is the unembedding matrix transforming the hidden states into scores over the vocabulary, $\circ$ is an element-wise product with broadcasting, and $\mathbf{x}^l_{-1}$ denotes the input of FFN in the last token before predicting $\hat{y}$ at $l$-th layer.
For a given threshold $\eta$, the key activated neurons for task sample $t=(x,y)$ is defined as: 

\begin{equation}
\resizebox{0.92\hsize}{!}{$
N_{\text{activated}}(t)
= \Bigl\{
(i, l)
\;\Big|\;
\exists \hat{y}_j \in y_i, f_{\text{neuron}}\left(i, l,\,\hat{y}_j \mid x \oplus \hat{y}_{<j}\right) > \eta 
\Bigr\},
$}
\label{eq:neuron_contribution_case}
\end{equation} 
Where: $\hat y_{<j} = (\hat{y}_1,\,\hat{y}_2,\,\ldots,\,\hat{y}_{j-1})$ denotes the partial response sequence before the $j$-th token $\hat{y}_j$, $l=1, 2, ...,L$ represents the layer index, and $n=1,2,...,N$ indicates the neuron index in each layer. Thereby, MUI in Eq. (\ref{eq:MUI}) can be revised as
\begin{equation}
\textbf{MUI}_\text{neuron}(t)=\frac{|N_\text{activated}(t)|}{NL}
\end{equation}

These definitions can be naturally extended to multiple samples $\mathcal{T} = \{t_1, t_2, \dots, t_K\}$ as:
\begin{align}
\textbf{MUI}_{\text{neuron}}(\mathcal{T})=\frac{|\bigcup_{i=1}^KN_\text{activated}(t_i)|}{NL}
\end{align}

\subsection{\ourmethod{}}
Using MUI as a guiding signal, our objective is to select a subset of $k$ representative samples $S = \{t_{i_1}, \ldots, t_{i_k}\}$ from the $K$-sized full dataset $\mathcal{T}=\{t_i\}_{i=1}^K$ (where $k < K$), such that the selected samples collectively maximize the coverage of the model’s capabilities. This objective is equivalent to maximizing the total MUI over the selected subset:

\begin{equation}
S=\arg\max_{S\subseteq\mathcal{T}} \textbf{MUI}_\text{neuron}(S)=\arg\max_{S\subseteq\mathcal{T}}|\bigcup_{t\in S}N_\text{activated}(t)|
\label{eq:optim_target}
\end{equation}

Given that the calculation of MUI depends on the specific model, we refer to the model used for this calculation as the indicator. Owing to shared capabilities across models, the resulting coverage has strong generalizability: the evaluated capabilities covered by the indicator are also tested when using the representative subset on other models. Detailed quantitative analysis and further discussion of this generalization are provided in the experiments.

\begin{algorithm}[htb]
\caption{MUI Data Selection}
\label{alg:algorithm}
\begin{flushleft}
\textbf{Input}: Key neuron set of original dataset $\{N_\text{activated}(t_i)\}_{i=1}^K$\\
\textbf{Parameter}: Subset size $k$\\
\textbf{Output}: Selected samples $S=\{t_{i_1},...,t_{i_k}\}$
\end{flushleft}
\begin{algorithmic}[1]
\STATE Initialize $S \leftarrow \emptyset$, $N_\text{covered} \leftarrow \emptyset$
\FOR{$t = 1$ \textbf{ to } $k$}
    \STATE Select $t_{i^*} = \arg\max_{t_i \notin S} |N_\text{covered} \cup N_\text{activated}(t_i)|$
    \STATE $S \leftarrow S \cup \{t_{i^*}\}$
    \STATE $N_\text{covered} \leftarrow N_\text{covered} \cup N_\text{activated}(t_{i^*})$
\ENDFOR
\STATE \textbf{return} $S$
\end{algorithmic}
\end{algorithm}

Given the optimization objective in Eq. (\ref{eq:optim_target}), this problem can be formulated as a Maximum Coverage Problem (MCP). Although NP-Hard, it admits an efficient greedy algorithm that iteratively selects the element providing the largest marginal gain. Despite its simplicity, the greedy approach with random sampling guarantees a $(1-1/e)$ approximation ratio~\cite{nemhauser1978analysis}. In this work, we adopt this method to solve the maximum coverage problem, as illustrated in Algorithm~\ref{alg:algorithm}. The overall process of \ourmethod{} is shown in Figure~\ref{fig:workflow}. Suppose we are given a full dataset $\mathcal{T} = \{q_1, q_2, q_3, q_4\}$. Each question $q_i \in \mathcal{T}$ is first mapped to its corresponding set of activated indicator neurons $N_\text{activated}(q_i)$, which represent the latent capabilities of the model triggered by that sample. For instance, $q_1$, $q_2$, and $q_4$ (from the network security domain) activate overlapping neurons, while $q_3$ (from the legal domain) activates a distinct region. Our goal is to select a subset $S$ that maximizes the total MUI, i.e., the union of activated neurons across selected samples. Under this objective, the subset $\{q_3, q_4\}$ achieves broader neuron coverage and thus better represents the model’s full capability spectrum.
\section{Experiments}
To comprehensively evaluate the effectiveness of our selected subsets as representatives of the full datasets, we conduct extensive experiments across four widely used benchmarks and 17 models, including both open-source and closed-source models. 
\begin{table*}[htp]
\small
\centering
\begin{tabular}{l|ccc|ccc|ccc|ccc}
\toprule
\multirow{2}{*}{Method} 
    & \multicolumn{3}{c|}{\textbf{GSM8K} ($k$ = 100)}
    & \multicolumn{3}{c|}{\textbf{ARC} ($k$ = 100)} 
    & \multicolumn{3}{c|}{\textbf{Hellaswag} ($k$ = 100)} 
    & \multicolumn{3}{c}{\textbf{MMLU} ($k$ = 100)} \\
&  $r_S$ & $r_K$ & MAE ↓ & $r_S$ & $r_K$ & MAE ↓ & $r_S$ & $r_K$ & MAE ↓ & $r_S$ & $r_K$ & MAE ↓ \\
\midrule
Random & \underline{95.3} & \underline{87.4} & 2.88 & 95.4 & 86.5 & 2.88 & 97.8 & 91.0 & 3.35 & 95.7 & 85.8 & 3.59 \\ 
K-Means & 95.0 & 87.0 & \underline{2.76} & \underline{95.8} & \underline{87.2} & \underline{2.78} & \underline{98.1} & \underline{91.5} & \underline{3.30} & 95.8 & 86.5 & 4.59 \\
tinyBenchmarks & 89.5 & 79.6 & \textbf{2.12} & 95.4 & 85.1 & 3.29 & \textbf{98.3} & 91.2 & 6.78 & \underline{96.8} & \underline{87.8} & \textbf{2.95} \\
\midrule
\ourmethod{} & \textbf{99.2} & \textbf{95.9} & 4.07 & \textbf{96.0} & \textbf{87.4} & \textbf{2.27} & \textbf{98.3}$^{\dagger}$ & \textbf{92.5}$^{\dagger}$ & \textbf{3.09}$^{\dagger}$ & \textbf{96.9} & \textbf{89.1} & \underline{3.45}\\
\bottomrule
\end{tabular}

\caption{Comparison between \ourmethod{} and tinyBenchmarks in terms of 1) correlation coefficient ($r_S, r_K$) between the evaluated model performances on the selected subset and those on the full dataset, and 2) MAE. Entries marked with a dagger ($\dagger$) indicate results obtained using \texttt{Qwen2.5-7B-Instruct}, which is used in place of LLaMA due to safety restrictions that significantly degrade LLaMA's performance. Our method outperforms the baselines with higher correlation coefficients and lower MAE. The best results are highlighted in bold, and the second-best results are underlined.}
\label{tab:tinybenchmark}
\end{table*}

\begin{table*}[htb]
\small
\centering
\begin{tabular}{l|ccc|ccc|ccc|ccc}
\toprule
\multirow{2}{*}{Method} 
    & \multicolumn{3}{c|}{\textbf{GSM8K} ($k$ = 237)}
    & \multicolumn{3}{c|}{\textbf{ARC} ($k$ = 145)} 
    & \multicolumn{3}{c|}{\textbf{Hellaswag} ($k$ = 93)} 
    & \multicolumn{3}{c}{\textbf{MMLU} ($k$ = 96)} \\
&  $r_S$ & $r_K$ & MAE ↓ & $r_S$ & $r_K$ & MAE ↓ & $r_S$ & $r_K$ & MAE ↓ & $r_S$ & $r_K$ & MAE ↓ \\
\midrule
Random & 97.4 & 91.3 & \textbf{1.75} & 95.4 & 85.8 & \textbf{2.17} & 97.8 & 91.1 & 3.41 & 95.5 & 85.7 & \underline{4.71}\\
K-Means & \underline{98.0} & \underline{91.4} & \textbf{1.75} & 95.5 & 85.8 & \underline{2.50} & \underline{98.2} & \underline{91.9} & 3.49 & 95.8 & 86.5 & 4.74\\
metabench & 97.0 & 89.3 & \underline{2.30} & \textbf{98.8} & \textbf{93.7} & 3.27 & 96.5 & 85.5 & \textbf{3.02} & \textbf{98.5} & \textbf{93.3} & 6.59\\
\midrule
\ourmethod{} & \textbf{98.5} & \textbf{93.7} & 3.62 & \underline{97.1} & \underline{87.8} & 2.77 & \textbf{98.4}$^{\dagger}$ & \textbf{92.8}$^{\dagger}$ & \underline{3.29}$^{\dagger}$ & \underline{97.1} & \underline{88.5} &  \textbf{3.63} \\
\bottomrule
\end{tabular}
\caption{Comparison between \ourmethod{} and metabench in terms of correlation coefficients ($r_S$, $r_K$) and MAE. The evaluation setting is consistent with Table~\ref{tab:tinybenchmark}, except that the subset size $k$ for each dataset is set to match the value used in \texttt{metabench}.}
\label{tab:metabench}
\end{table*}
\subsection{Experiment Setting}
\begin{description}[style=unboxed, leftmargin=0cm]
\item[\textbf{Baselines.}] We compare the following baselines:
  \begin{itemize}
      \item \textbf{Random Selection}, where the representative datasets are randomly selected from the full dataset. 
      \item \textbf{Representation-Based Clustering}, where representative samples are selected based on clustering results. Specifically, we compute question representation embeddings using \texttt{text-embedding-3-large}, and then perform $k$-means clustering to group the questions, following~\cite{pacchiardi2024100} (marked as K-Means).
      \item \textbf{SOTA methods}:
  for other publicly available selection methods, such as tinyBenchmarks~\cite{polo2024tinybenchmarks} and metabench~\cite{kipnis2024metabench}, we directly compare our method with their released representative subsets\footnote{\url{https://huggingface.co/tinyBenchmarks}}\footnote{\url{https://huggingface.co/datasets/HCAI/metabench}}.
  \end{itemize} 
  \item[\textbf{Datasets.}] To comprehensively evaluate the applicability, following~\cite{polo2024tinybenchmarks, kipnis2024metabench, yingllms}, we selected four benchmarks covering diverse domains: GSM8K~\cite{cobbe2021training} for math reasoning; ARC-Challenge~\cite{clark2018think} for scientific reasoning; Hellaswag~\cite{zellers2019hellaswag} for commonsense inference and MMLU~\cite{hendrycks2020measuring} for general tasks. The detailed statistics result is shown in the Appendix.
  
  \item[\textbf{Comparison Models.}] 
  To thoroughly evaluate our method, we select 17 models of varying scales from diverse sources, including both open-source and closed-source models. The selection covers both models with significant capability gaps and those with similar performance to simulate real-world scenarios. We assess the data sampling method by comparing the relative performance of these models on the original dataset versus the sampled subsets.
  The evaluated models include: 
  (1) Qwen series~\cite{qwen,qwen2.5,yang2024qwen25mathtechnicalreportmathematical,qwen3technicalreport}; 
  (2) LLaMA series~\cite{vicuna2023,touvron2023llama,grattafiori2024llama,guo2025deepseek};
  (3) Gemma series~\cite{gemma_2024};
  (4) proprietary models~\cite{openai2024gpt4o,google2025gemini25}.
  See Appendix for detailed model list and generation settings.
  
  \item[\textbf{Metrics.}] 
  To evaluate how well the selected subsets represent the full datasets, we consider several metrics to quantify the discrepancy between the subset and the original data. Specifically, we report \textbf{Spearman’s correlation} ($r_S$) and \textbf{Kendall’s $\tau$} ($r_K$) between model performances derived from the selected subset and those from the full dataset (i.e. relative ranking), following previous work~\cite{vivek2023anchor,polo2024tinybenchmarks}. In addition, we measure prediction fidelity using the \textbf{Mean Absolute Error (MAE)} between model performance on the subset and on the full dataset (i.e. absolute performance). We use the correlation coefficient as the primary metric, as discussed in the Introduction. Following~\cite{polo2024tinybenchmarks}, for methods involving randomness (Random, K-Means and \ourmethod{}), we repeat the selection methods $t=5$ times and average the above metrics results to ensure robustness.
\end{description}
\textbf{Implementation details} 
For the threshold $\eta$ of MUI in Eq. (\ref{eq:neuron_contribution_case}), we adopt a layer-wise top-$k$\% threshold following~\cite{cao2025model}. Detailed implementation details are provided in the Appendix.
For the selection of $k$ in Algorithm~\ref{alg:algorithm}, we set the subset size to match the sizes of publicly available representative datasets to ensure fair comparison with baselines. Since existing representative subsets may not fully cover the evaluation capabilities of the full benchmark
, we also consider an adaptive coverage-based $k$ setting: that is, we select the smallest data size that achieves a predefined coverage threshold of $r = 0.8$, formally:
\begin{equation}
k^*=\min_k\{k, |N_\text{covered}(k)|/|N_\text{covered}(K)| \ge r\}
\label{eq:optim_k}
\end{equation}
Unless specified, we choose LLaMA3.1-8B-Instruct as the indicator model. For more experiments using more different models, please refer to Ablation Study.

\subsection{Effectiveness of \ourmethod{}}
For a fair comparison, we follow the settings of SOTA methods and sample the same number of evaluation samples from each original benchmark as they do. Note that this number may not be optimal for our method, and we will discuss the issue of subset size in later sections.
The correlation results across the 17 selected models shown in Table~\ref{tab:tinybenchmark} and Table~\ref{tab:metabench} indicate that:
1) Our method achieves strong relative correlation across all datasets, with low correlation variance (please refer to the Appendix for details), demonstrating the effectiveness of our approach;
2) In the ARC and MMLU datasets shown in Table~\ref{tab:metabench}, our selection process exhibits slight sensitivity to the subset size and does not achieve optimal results. This may be because, when the number of samples is relatively small in these challenging datasets, the model’s capabilities cannot be consistently and fully covered, leading to certain fluctuations.
3) On the MAE metric, the results vary significantly across different methods, suggesting that it may not be a reliable indicator --- consistent with our earlier discussion on relative versus absolute performance scores;
4) Compared to $r_K$, the $r_S$ metric is generally less sensitive to pairwise ranking errors, leading to consistently higher scores across different methods. In contrast, $r_K$ provides a more fine-grained characterization of how well the sampling method preserves model rankings, and its values are generally lower. For example, on certain datasets such as GSM8K, even SOTA methods achieve only 89.3\%.
5) The K-Means baseline does not show a significant advantage over the random baseline, yet it incurs higher computational overhead. This demonstrates that the random strategy is already a strong baseline. However, the effectiveness of the random method may be overestimated, as repeated sampling effectively increases the actual size of the subset. For the variability comparison, please refer to the Appendix.
\subsection{Capability Coverage}
\begin{figure*}[htb]
  \centering
  \includegraphics[width=0.95\textwidth,page=1]{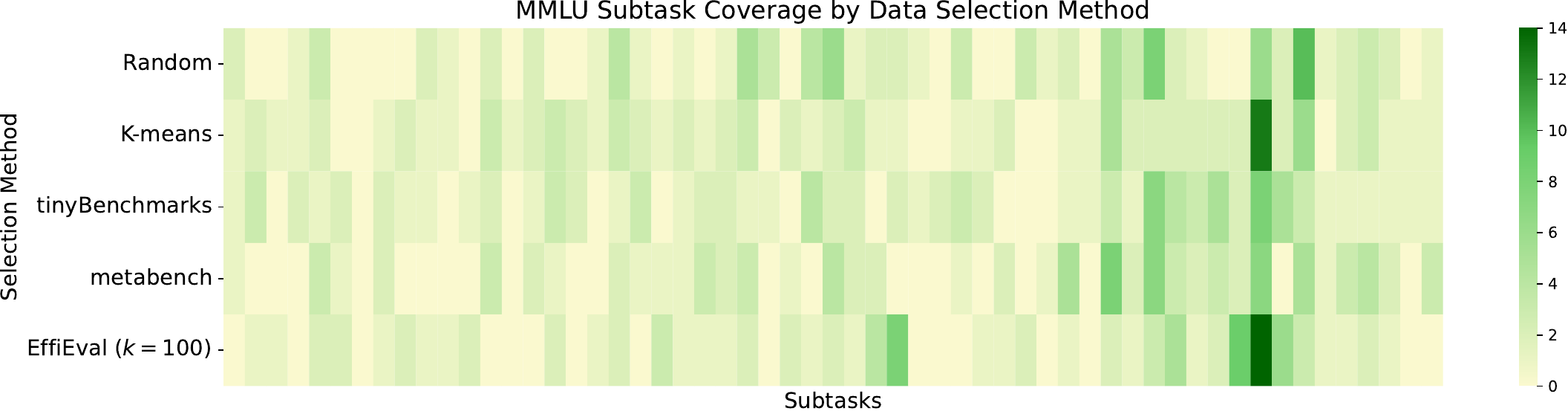}
  \caption{Category coverage of the representative dataset baselines and \textbf{\ourmethod{}} (with sample number set to fixed $k=100$) on MMLU (57 categories in total). Each cell's color intensity reflects how many samples are selected in the corresponding category. The heatmap illustrates that previous methods do not fully cover the categories of the dataset, indicating that the subset size $k$ of these baselines may be improper. The horizontal axis corresponds to the 57 category labels of the MMLU dataset, while the vertical axis lists the different selection methods.}
  \label{fig:baseline_cover}
\end{figure*}

SOTA methods typically select fixed, and small benchmark subsets (e.g., tinyBenchmarks with only 100 samples). We argue that the fixed size of these subsets makes it difficult to dynamically adjust their size, and that their selection is often insufficient to capture the full dataset’s content coverage. 
In our in-depth analysis, we quantitatively examine the representativeness issues arising from such limited data. For example, as shown in Figure~\ref{fig:baseline_cover}, the 100 representative samples selected by tinyBenchmarks and metabench cover only 46 and 37 out of 57 sub-tasks in the MMLU benchmark, respectively. From the perspective of model capability activation, the MUI activated by these subsets accounts for approximately 10\% of that activated by the full dataset for the indicator model (see Appendix for more details). These observations suggest that \textbf{such SOTA methods fail to select a representative subset with both appropriate size and distribution to ensure adequate coverage of model capabilities}.
In contrast, selecting data based on MUI capability coverage can readily address the insufficient capability coverage of existing methods. Instead of searching for an appropriate subset size $k$ for each target benchmark, we iteratively add data samples to the subset $S$ using \ourmethod{} until the coverage ratio reaches a threshold $r$. 
For example, when $r = 0.8$ is set to ensure high data quality, our method achieves complete category coverage on the MMLU dataset, covering all 57 subtasks. Importantly, this process allows the threshold to be flexibly adjusted based on efficiency requirements.
Furthermore, from the perspective of effectively substituting for the original dataset, the results in Table~\ref{tab:dynamic_k} show that subsets selected by \ourmethod{} demonstrate clear advantages over both random selection and K-means clustering approaches, yielding superior performance. \textbf{Overall, these results highlight the effectiveness and scalability of our approach.}

\begin{figure*}[htb]
  \centering
  \begin{subfigure}[b]{0.24\textwidth}
    \includegraphics[width=\textwidth]{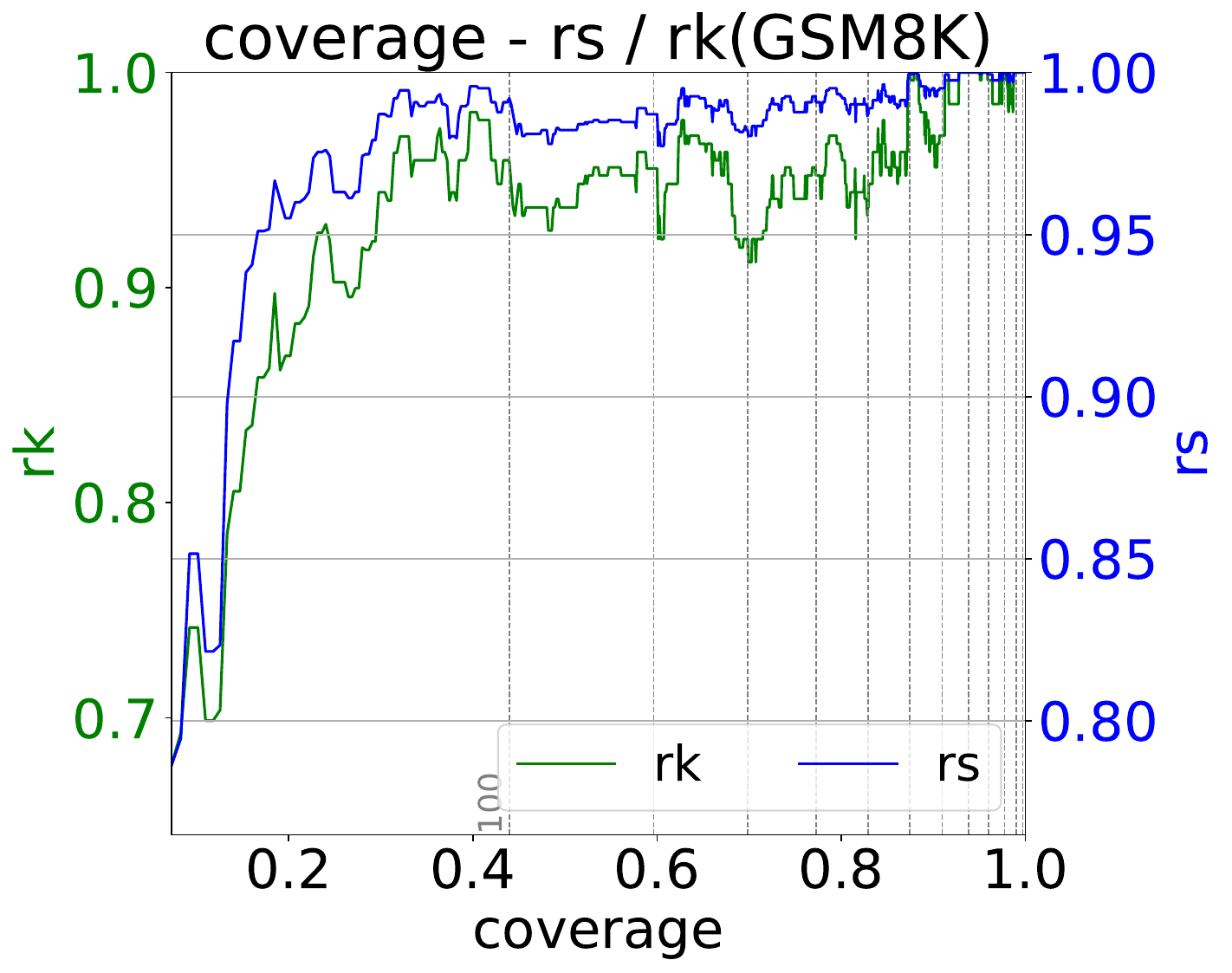}

  \end{subfigure}
  \begin{subfigure}[b]{0.24\textwidth}
    \includegraphics[width=\textwidth]{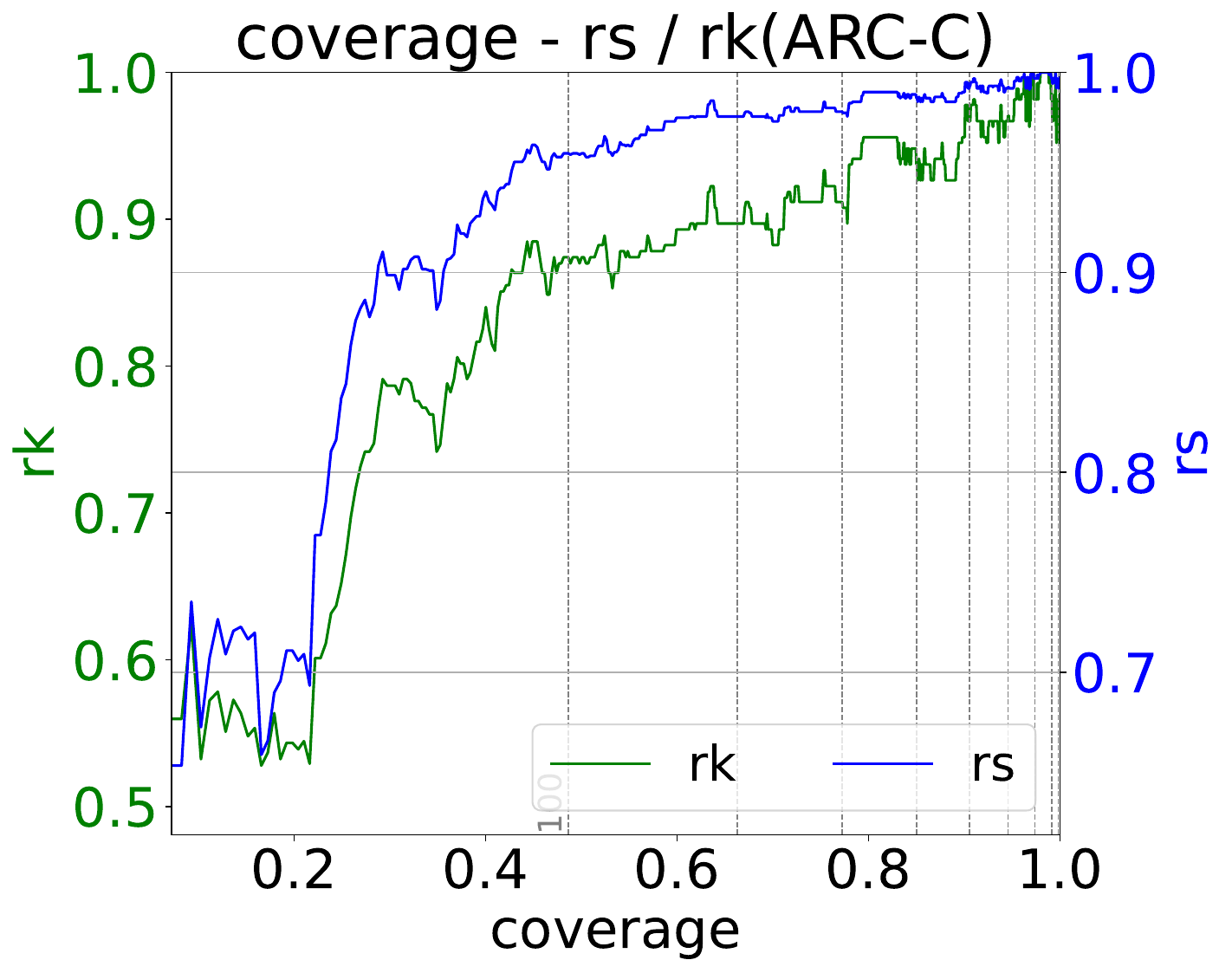}

  \end{subfigure}
  \begin{subfigure}[b]{0.24\textwidth}
    \includegraphics[width=\textwidth]{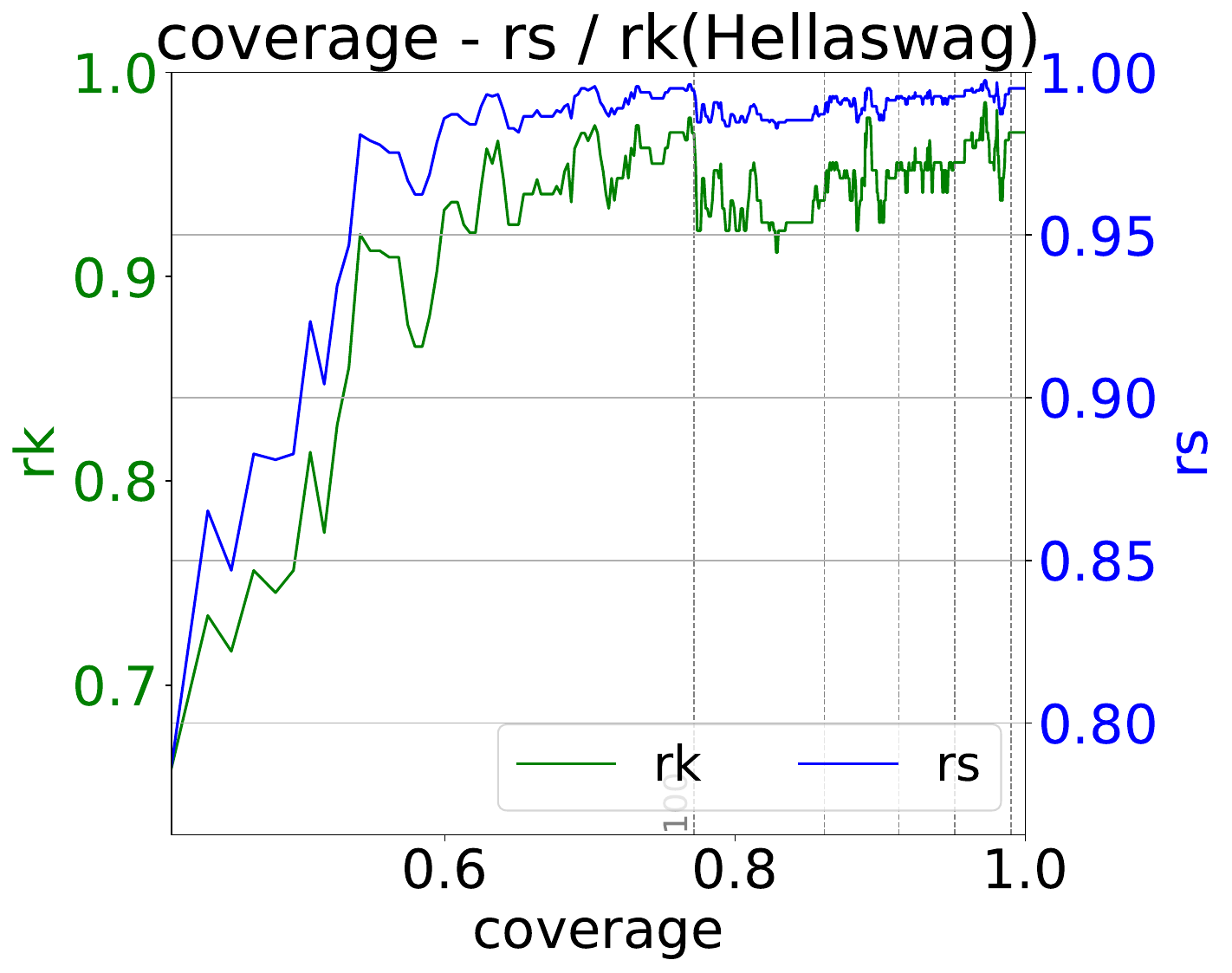}

  \end{subfigure}
  \begin{subfigure}[b]{0.24\textwidth}
    \includegraphics[width=\textwidth]{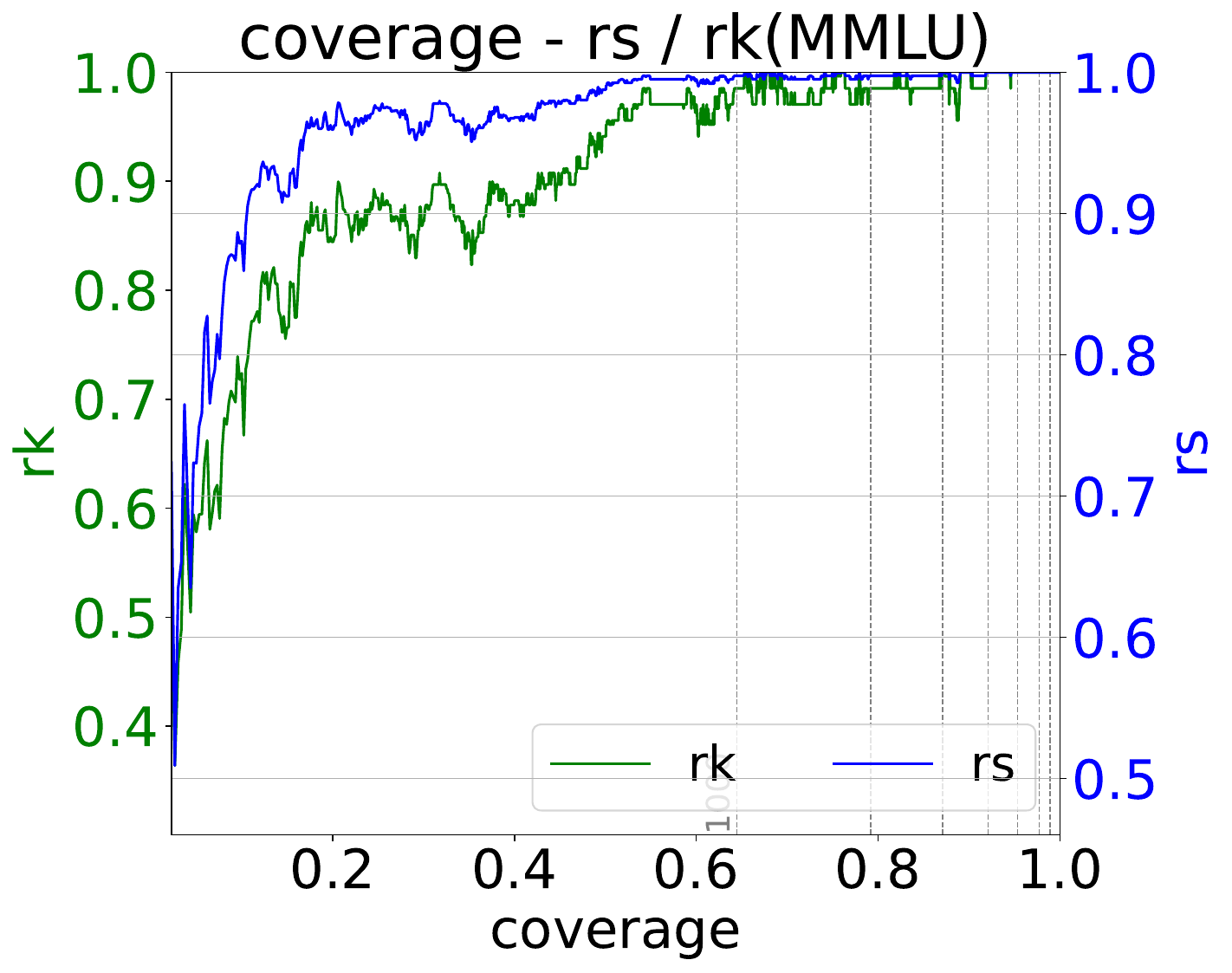}

  \end{subfigure}
  \caption{A dual-axis plot of $r_K$ (green) and $r_S$ (blue) versus coverage on four benchmarks. Gray vertical dashed lines mark the coverage levels corresponding to evenly spaced values of sample size $k$, with the step size indicated at the first line. The figure illustrates how much data is needed for different datasets to achieve a trade-off between correlation and evaluation efficiency.}
  \label{fig:coverage_r}
\end{figure*}

\subsection{Determine Subset Size Adaptively}

\begin{table}[htb]
\centering
\small
\setlength{\tabcolsep}{4pt} 
\begin{tabular}{lccccc}
\toprule
Method & Dataset & $k \ (k/K)$ & $r_S$ & $r_K$ & MAE ↓\\
\midrule
Random & \multirow{3}{*}{GSM8K}& \multirow{3}{*}{446 (33.8\%)} & 98.9 & 94.6 & \textbf{1.03} \\
K-means & & & \underline{99.2} & \underline{96.1} & \underline{1.40} \\
\ourmethod{} & & & \textbf{99.3} & \textbf{96.3} & 1.99\\
\midrule
Random & \multirow{3}{*}{ARC}& \multirow{3}{*}{332 (28.3\%)}  & 98.6 & 93.8 & \textbf{1.47} \\
K-means & & & \underline{98.7} & \underline{94.3} & 1.52 \\
\ourmethod{} & & & \textbf{99.0} & \textbf{95.6} & \underline{1.48}\\
\midrule
Random & \multirow{3}{*}{Hellaswag}& \multirow{3}{*}{125 (1.2\%)}  & 97.9 & 91.1 & 3.18 \\
K-means & & & \underline{98.2} & \underline{92.0} & \underline{3.11} \\
\ourmethod{} & & & \textbf{98.5}$^{\dagger}$ & \textbf{92.3}$^{\dagger}$ & \textbf{2.78}$^{\dagger}$ \\
\midrule
Random & \multirow{3}{*}{MMLU}& \multirow{3}{*}{2082 (14.8\%)}  & 99.4 & 96.8 & \textbf{0.94} \\
K-means & & & \underline{99.7} & \underline{98.0} & \underline{1.97} \\
\ourmethod{} & & & \textbf{99.8} & \textbf{98.5} & \textbf{0.94} \\
\bottomrule
\end{tabular}
\caption{Effectiveness of the representative subset based on a coverage ratio $r=0.8$. The efficiency is quantified by the ratio of the selected subset size to the full dataset size ($k/K$).}
\label{tab:dynamic_k}
\end{table}
By selecting data to achieve a coverage ratio of $r = 0.8$, we have already demonstrated that the resulting subsets can effectively serve as substitutes for the full dataset. In this section, we further investigate the relationship between different coverage thresholds and the effectiveness of the selected subsets. To investigate this, we evaluate the effectiveness of representative subsets selected under different coverage thresholds, following the experimental settings in Experiment Setting Section. The result shown in Figure~\ref{fig:coverage_r} indicates that when $r$ is small, the selected subset fails to sufficiently represent the full dataset, leading to low correlations. As $r$ increases, the correlation improves, finally converges to 1. Moreover, Figure~\ref{fig:coverage_r} reveals large variation in information density across benchmarks. For example, on Hellaswag, selecting just 125 samples (1\% of the original dataset) covers over 80\% of the neurons activated by the full dataset, whereas MMLU requires approximately 2000 samples (14.8\% of the original dataset) to reach the same coverage.
Notably, on certain benchmarks (e.g., MMLU), high correlations can be achieved even at lower coverage thresholds (e.g., $r=0.6$). We hypothesize that this is because some capabilities --- though controlled by different neurons --- are highly interrelated. As a result, \textbf{it is possible to adaptively select an efficient coverage threshold based on the characteristics of different datasets, while still maintaining high evaluation effectiveness. This highlights the flexibility and adaptability of our approach.} To select a value of $k$ that balances evaluation efficiency and reliability, we search across all possible $k$ values for a window in which the correlation remains both high and stable. Specifically, we identify the earliest window of 200 consecutive $k$ values that satisfies the following two conditions:
1) Reliability: Each $k$ in the window yields a Kendall’s $\tau$ correlation above 0.9;
2) Stability: The correlation values within the window are stable, meaning their Kendall’s $\tau$\footnote{Note that this Kendall’s $\tau$ measures the stability of the correlation sequence, and is different from the one used as our primary evaluation metric.} with respect to a monotonically increasing index is less than 0.1 in absolute value.
The midpoint of this window is then selected as the final value of $k$.
Table~\ref{tab:ablation_k} summarizes the evaluation metrics under this setting of $k$ values. Compared with fixed-$r$ selection, this strategy further reduces the subset size—by up to 95\% --- without sacrificing too much correlation.

\begin{table}[htbp]
\centering
\small
\setlength{\tabcolsep}{4pt} 
\begin{tabular}{lccccc}
\toprule
Method & Dataset & $k \ (k/K)$ & $r_S$ & $r_K$ & MAE ↓\\
\midrule
Random & \multirow{3}{*}{GSM8K}& \multirow{3}{*}{140 (10.6\%)} & 96.3 & 89.5 & \textbf{2.37} \\
K-means & & & \underline{96.6} & \underline{89.7} & \underline{2.44} \\
\ourmethod{} & & & \textbf{98.2} & \textbf{93.7} & 4.38 \\
\midrule
Random & \multirow{3}{*}{ARC}& \multirow{3}{*}{409 (34.9\%)}  & 98.6 & 93.8 & \underline{1.33} \\
K-means & & & \textbf{99.3} & \textbf{96.6} & \textbf{1.08} \\
\ourmethod{} & & & \underline{98.7} & \underline{94.1} & 1.60\\
\midrule
Random & \multirow{3}{*}{Hellaswag}& \multirow{3}{*}{149 (1.5\%)}  & \underline{98.2} & 91.9 & \textbf{2.72} \\
K-means & & & \underline{98.2} & \underline{92.2} & \underline{2.76} \\
\ourmethod{} & & & \textbf{98.5}$^{\dagger}$ & \textbf{92.6}$^{\dagger}$ & 3.32$^{\dagger}$ \\
\midrule
Random & \multirow{3}{*}{MMLU}& \multirow{3}{*}{669 (4.8\%)}  & 98.8 & 95.1 & 1.88 \\
K-means & & & \underline{99.1} & \underline{95.3} & \underline{1.63} \\
\ourmethod{} & & & \textbf{99.5} & \textbf{97.1} & \textbf{1.55} \\
\bottomrule
\end{tabular}
\caption{Effectiveness of the representative subset based on stable sliding window.}
\label{tab:ablation_k}
\end{table}

\subsection{Ablation Study}

\begin{description}[style=unboxed, leftmargin=0cm]
  \item[Independence between \ourmethod{} and performance.] A good selection process should be orthogonal to the models' performance, otherwise it could introduce bias. To verify that \ourmethod{} is performance-independent, we repeatedly sample 10\% of the smaller group between the correct and incorrect samples from four benchmarks, and repeat this process 10 times, then compute MUI of the indicator model on the two sets, finally conduct Mann-Whitney U test to verify that there is no statistically significant difference between the two distributions. As shown in Figure~\ref{fig:independence}, no significant difference is observed across performance-based groupings, demonstrating that MUI is independent of model performance. This property guarantees that \ourmethod{} selects representative samples in a performance-agnostic manner.
\begin{figure}[htp]
  \centering
  \includegraphics[width=0.45\textwidth,page=1]{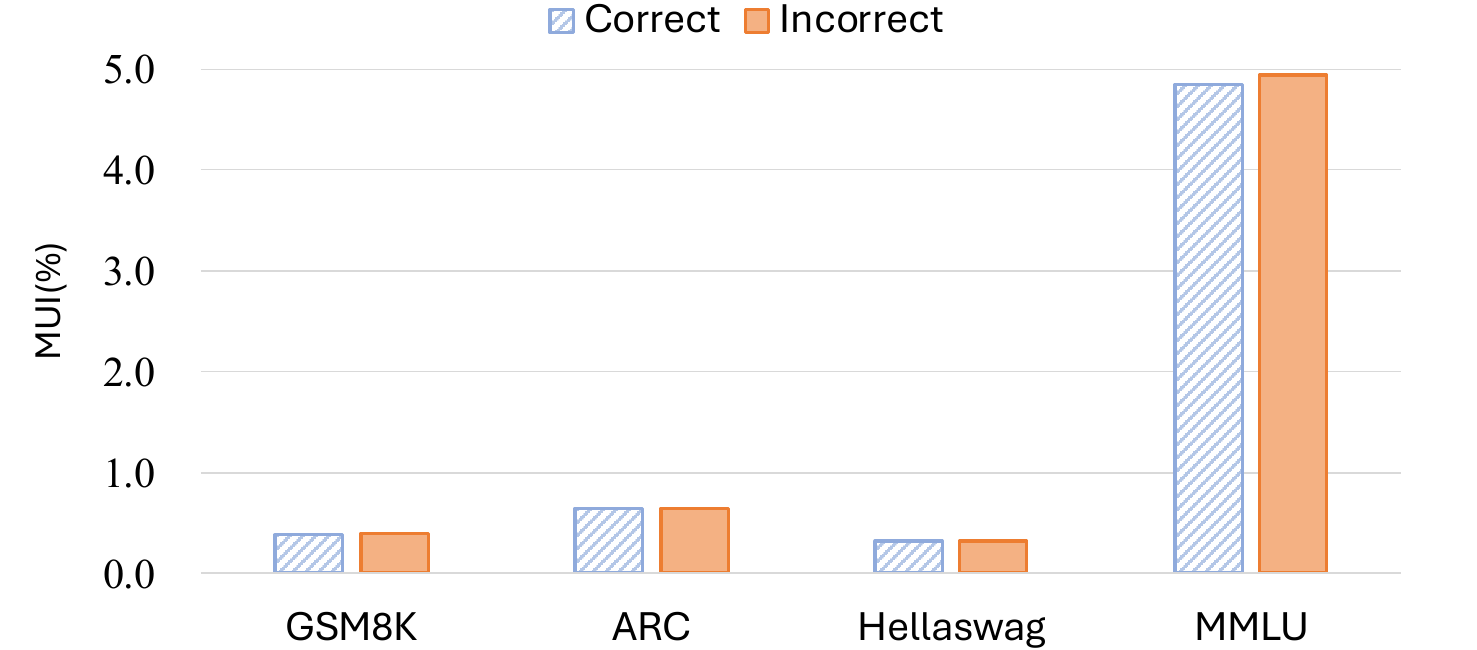}
  \caption{MUI of correct samples and incorrect samples on four benchmarks. The Mann-Whitney U test fails to reject the null hypothesis, suggesting that there is no significant difference between the two distributions.}
  \label{fig:independence}
\end{figure}

  \item[Generalization of the indicator model.] In the experiments above, we mainly use LLaMA-3.1-8B-Instruct to identify key neurons, suggesting that a single model is sufficient to select a representative subset that covers the capabilities of other models. We hypothesize that this may be due to a high degree of correlation in capability distributions across different models. To further investigate the generalization of the indicator model, we explore how much the choice of model affects the selection outcome. For this purpose, we conduct additional experiments on selected benchmarks with Qwen2.5-7B-Instruct and Qwen-1.5-7B-Chat, with $k$ set identical to those in Table~\ref{tab:dynamic_k}. The experiment results are shown in Table~\ref{tab:ablation_model}. In most cases, replacing the indicator model still successfully guides the data selection process, achieving strong performance correlations. 
  For the Qwen series on the MMLU benchmark, some of the selected questions are relatively simple, making it difficult to distinguish between models and resulting in low correlation scores. This observation highlights the direction for our future work.
\begin{table}[htb]
\small
\setlength{\tabcolsep}{5pt} 
\centering
\begin{tabular}{l|ccc|ccc}
\toprule
\multirow{2}{*}{Method} 
    & \multicolumn{3}{c|}{\textbf{GSM8K} ($k$ = 446)}
    & \multicolumn{3}{c}{\textbf{MMLU} ($k$ = 2082)} \\
&  $r_S$ & $r_K$ & MAE ↓ & $r_S$ & $r_K$ & MAE ↓ \\
\midrule
Random & 98.9 & 94.6 & 1.03  & 99.4 & 96.8 & 0.94 \\
K-Means & 99.2 & 96.1 & 1.40 & 99.7 & 98.0 & 1.97 \\
\midrule
LLaMA3.1-8B & 99.3 & 96.3 & 1.99  & 99.8 & 98.5 & 0.94 \\ 
Qwen2.5-7B & 98.9 & 94.7 & 2.32 & 96.9 & 88.8 & 10.85\\
Qwen1.5-7B & 99.1 & 95.2 & 1.09 & 97.9 & 91.2 & 9.41 \\
\bottomrule
\end{tabular}
\caption{Comparison between different indicator models. The selection remains effective when switching the indicator model, showing strong performance correlations and supporting the generalization of our method. Results on other benchmarks are provided in the Appendix.}
\label{tab:ablation_model}
\end{table}
\end{description}

\section{Discussion}
Despite the effectiveness of our method, several limitations remain and deserve further exploration. The correlation coefficient may not serve as an ideal proxy for thoroughly evaluating the representativeness of data. Some datasets repeatedly test the same underlying capabilities of a model. When sampling based on capability coverage, the resulting subset may distort the original distribution, thereby weakening the performance correlation between the subset and the full dataset. In such cases of data redundancy, preserving the original model ranking may no longer be the most appropriate objective. Future work should explore more reasonable and effective evaluation strategies that account for redundancy, diversity, and the actual capabilities being tested.
\section{Conclusion}
In this paper, we address the urgent need for efficient and reliable evaluation in the era of LLMs. We propose \ourmethod{}, a training-free approach for benchmark subset selection that maximizes internal capability coverage as measured by the MUI. Our method is specifically designed to satisfy three key criteria for high-quality evaluation: representativeness, fairness, and generalizability. Extensive experiments on multiple public benchmarks demonstrate that \ourmethod{} consistently achieves strong correlation with full-dataset evaluation. Moreover, our approach is flexible and scalable, enabling users to adjust the trade-off between evaluation efficiency and coverage. In the future, we plan to further explore how to better control the quality of selected subsets, and to continuously improve our approach as advances in interpretability and model analysis emerge.
\bibliography{aaai2026}

\newpage
\appendix
\section{Generation Settings}
In all experiments, we adopt consistent decoding settings to ensure fair comparison across models. Specifically, we set the maximum number of newly generated tokens (\texttt{max\_new\_tokens}) to 8192 for reasoning models (DeepSeek-R1-Distill-Qwen-7B, DeepSeek-R1-Distill-LLaMA-8B and Qwen-3-8B with thinking enabled), and 1024 for all other models. We use top-p sampling with \texttt{top\_p} = 1.0, and fix the temperature to 0.0 to disable sampling and enforce deterministic generation.

\section{Implementation for Threshold Function}
\label{sec:threshold}
In this paper, we adopt a layer-wise top-$k$\% threshold function for key neuron selection, i.e.
$$
\eta(l,topk)=V_l^{\lfloor N\cdot topk\rfloor}
$$
Where $V_l$ is the sorted (in descending order) activation scores on all neurons and in output tokens in layer $l$, $topk\in(0,1)$ is the threshold hyperparameter, $N$ is the neuron number per layer. In all experimental settings, we set $topk$ = 0.1\%. This approach can alleviate discrepancies due to different model architectures when calculating the MUI~\cite{cao2025model}.

\section{Case Study}

In this section, we illustrate the execution process of \ourmethod{} to better demonstrate how the algorithm selects samples with previously uncovered capabilities. For each candidate sample, we present its most similar counterpart (i.e., a sample whose capabilities have already been covered) and its most dissimilar one (i.e., the sample most likely to be selected by \ourmethod{} in the next step). The similarity metric is defined as the Jaccard distance between the sets of activated neurons:
$$\text{dist}(t_i,t_j)=\text{JaccardDistance}(N_\text{activated}(t_i), N_\text{activated}(t_j))$$

\begin{tcolorbox}[grayboxstyle, title=Case Study on GSM8K]
\textbf{Prompt:}

Digimon had its 20th anniversary.  When it came out John was twice as old as Jim.  If John is 28 now how old is Jim? 

\textbf{(One-variable linear equation)}

\textcolor[rgb]{0,0.8,0}{\textbf{Most similar prompt:}}

Liam is 16 years old now. Two years ago, Liam’s age was twice the age of Vince. How old is Vince now? 

\textbf{(One-variable linear equation)}

\textcolor[rgb]{0.8,0,0}{\textbf{Most dissimilar prompt:}}

My kitchen floor has a total area of 200 SqFt. I want to install new square floor tiles that cost \$12 each, and each tile side is 1ft in length. How much will it cost me to renovate my kitchen floor? 

\textbf{(Area Computation)}
\end{tcolorbox}

\begin{tcolorbox}[grayboxstyle, title=Case Study on ARC-Challenge]
\textbf{Prompt:}

A student pushed a large rubber ball on a flat, frictionless surface. The ball rolled at a speed of 1 meter per second. Which statement best describes the motion of the ball when the student stopped pushing the ball?

A. The ball accelerated.

B. The ball did not move.

C. The ball changed direction.

D. The ball continued to move in the same direction.

\textbf{(Physics)}

\textcolor[rgb]{0,0.8,0}{\textbf{Most similar prompt:}}

Two girls are pulling on opposite ends of a thick rope. Both girls pull on the rope with the same force but in opposite directions. If both girls continue to pull with the same force, what will most likely happen? 

A. One girl will pull the other toward her.

B. Both girls will stay in the same place.

C. Gravity will cause the rope to sag.

D. The rope will break.

\textbf{(Physics)}

\textcolor[rgb]{0.8,0,0}{\textbf{Most dissimilar prompt:}}

Which question about tulips could best be answered by scientific research? 

A. Are tulips better than other flowers?

B. What genes determine tulip petal color?

C. Why do people like to look at tulips?

D. Which color of tulips is the prettiest?

\textbf{(Biology)}
\end{tcolorbox}

\begin{tcolorbox}[grayboxstyle, title=Case Study on Hellaswag]
\textbf{Prompt:}

[header] How to use and install a live cd of linux [title] Make sure your computer is booting from the cd drive. [substeps] Either turn on or restart your computer. While doing this, hold the delete button to enter the bios.

1. [title] Once you've backed up your computer, press a. [step] After a few seconds the bios should pop up on your screen.

2. If you don't hear any result after a few seconds of working (or if the cd drive isn't booting freely from the computer), you may need to reboot your computer. [title] Run the cd containing the bios.

3. Use your left and right arrow keys to navigate to the boot tab. * once on the boot tab use your down arrow keys to navigate to the ``boot device priority'' menu.

4. The program is now on the cd drive and it cannot be added to the system at any time. To quit the program via the bios), enter the status key and select restart the computer.

\textbf{(Operating System)}

\textcolor[rgb]{0,0.8,0}{\textbf{Most similar prompt:}}

[header] How to reformat windows 7 [title] Backup all your files, drivers and settings so that you can restore them later. [title] Find all your installation discs or product keys for the programs you want to keep so that you can restore them after the installation is complete. [title] Partition your hard drive.

1. [step] For windows 7 you will need to partition all of your data, but this is optional as your computer needs your hard drive. [substeps] Right-click your drive and select partition all.

2. [step] Partition your hard drive to remove your usb storage device from cd or dvd. [title] Finish the process using back up documentation if you wish to keep a backup device.

3. [step] This means dividing the hard drive into parts and making the parts available to the os (operating system). [title] Click on ``start'' and then control panel.

4. [step] This helps to prevent any lost drives. You can also patch any issues that you have with windows 7.

\textbf{(Operating System)}

\textcolor[rgb]{0.8,0,0}{\textbf{Most dissimilar prompt:}}

A woman is bent over holding a weight bar. She picks the weight up and holds it at her shoulders. she

1. then pushes off the bar.

2. bends down and begins exercising using the weight bar.

3. then lifts the weight over her head.

4. lifts it to her chest and works out.

\textbf{(Commonsense Understanding)}
\end{tcolorbox}

\begin{tcolorbox}[grayboxstyle, title=Case Study on MMLU]
\textbf{Prompt:}

In the absence of intervention, imperfect competition, externalities, public goods, and imperfect information all result in which of the following?

A. Demand curves that should be added vertically

B. Market failure

C. Prices that are too low

D. Quantities of output that are too high

\textbf{(High School Microeconomics)}

\textcolor[rgb]{0,0.8,0}{\textbf{Most similar prompt:}}

A negative externality in the market for a good exists when

A. the market overallocates resources to the production of this good.

B. spillover benefits are received by society.

C. the marginal social benefit equals the marginal social cost.

D. total welfare is maximized.

\textbf{(High School Microeconomics)}

\textcolor[rgb]{0.8,0,0}{\textbf{Most dissimilar prompt:}}

In the current year Vinton exchanged unimproved land for an apartment building. The land had a basis of \$300000 and a fair market value (FMV) of \$420000 and was encumbered by a \$100000 mortgage. The apartment building had an FMV of \$550000 and was encumbered by a \$230000 mortgage. Each party assumed the other's mortgage. What is Vinton's basis in the office building?

A. \$300,000

B. \$320,000

C. \$430,000

D. \$550,000

\textbf{(Professional Accounting)}
\end{tcolorbox}

\section{Variability Analysis}
In this section, we verify the variability of the selection process using the Generalized Jaccard Index, which is defined as
$$
J(S_1,S_2,...,S_t)=\frac{|\bigcap_{i=1}^tS_i|}{|\bigcup_{i=1}^tS_i|}
$$
where $S_i$ denotes the sample index set obtained from the i-th run of the random selection process.
A lower value of this metric indicates that the selected subsets vary more across different runs, suggesting that the selection process is less stable. Conversely, a higher value implies greater stability and consistency in the selected samples. We fix $k=100$ and repeat the random selection process $t=5$ times. The results are shown in Table~\ref{tab:jaccard}. As indicated by the Generalized Jaccard Index, our method consistently selects highly overlapping subsets, even when the subset size is small, demonstrating a high degree of stability.

\begin{table}[htbp]
\centering
\begin{tabular}{lcccc}
\toprule
Method
    & \textbf{GSM8K}
    & \textbf{ARC}
    & \textbf{Hellaswag} 
    & \textbf{MMLU} \\
\midrule
Random & 0.0 & 0.0 & 0.0 & 0.0 \\
K-Means & \underline{1.1} & \underline{1.5} & \textbf{6.1} & \underline{8.0}\\
\ourmethod{} & \textbf{60.5} & \textbf{79.6} & \underline{5.9} & \textbf{44.5}\\
\bottomrule
\end{tabular}
\caption{Generalized Jaccard Index of different selection methods across $t=5$ runs on four benchmarks ($k=100$). A higher Jaccard Index reflects greater stability in the selected subsets.}
\label{tab:jaccard}
\end{table}

\section{Separability Analysis}
In this section, we assess the effectiveness of various data sampling strategies in distinguishing model performance using the Separability with Confidence metric~\cite{li2024crowdsourced}. This metric quantifies how confidently a benchmark can separate different models by computing the proportion of model pairs whose performance intervals do not overlap under repeated evaluations. Specifically, we apply each sampling method multiple times to generate subsets of evaluation data, and measure how often the subset can reliably identify a performance difference between two models. To obtain reliable percentile estimates for the confidence intervals, we set the number of sampling repetitions to $t=50$. A higher separability score indicates that the sampling method tends to produce subsets that preserve meaningful model distinctions, providing stronger signals for model comparison. As shown in Table~\ref{tab:separability}, our method achieves a higher separability score compared to random baselines, indicating a more stable and confident distinction between model performances. This suggests that our selected subsets tend to yield more consistent model rankings across repeated evaluations.

\begin{table}[htbp]
\centering
\begin{tabular}{lcccc}
\toprule
Method
    & \textbf{GSM8K}
    & \textbf{ARC}
    & \textbf{Hellaswag} 
    & \textbf{MMLU} \\
\midrule
Random & 55.2 & 52.2 & 68.4 & 44.9 \\
K-Means & \underline{60.3} & \underline{55.2} & \textbf{77.9} & \underline{55.2} \\
\ourmethod{} & \textbf{83.8} & \textbf{89.0} & \underline{73.5} & \textbf{82.4}\\
\bottomrule
\end{tabular}
\caption{Separability with Confidence scores of different sampling methods across four benchmarks ($k=100$, confidence level=95\%). Our method outperforms the random baseline on all benchmarks and achieves significantly higher separability than K-Means on three out of four benchmarks, with one benchmark showing slightly lower performance than K-Means. Overall, this demonstrates the strong and stable capability of our method to distinguish model performances across diverse tasks.}
\label{tab:separability}
\end{table}

\section{Computing Infrastructure}
All experiments were conducted on a server running Ubuntu 22.04.5 LTS, equipped with an Intel(R) Xeon(R) Platinum 8480+ CPU, 2TB of RAM, and a 96GB NVIDIA H20 GPU.

\begin{table*}[htbp]
\centering
\begin{tabular}{l|cc|cc|cc|cc}
\toprule
\multirow{2}{*}{Method} 
    & \multicolumn{2}{c|}{\textbf{GSM8K} ($k$ = 100)}
    & \multicolumn{2}{c|}{\textbf{ARC} ($k$ = 100)} 
    & \multicolumn{2}{c|}{\textbf{Hellaswag} ($k$ = 100)} 
    & \multicolumn{2}{c}{\textbf{MMLU} ($k$ = 100)} \\
&  $\text{std.}(r_S)$ ↓ & $\text{std.}(r_K)$ ↓ & $\text{std.}(r_S)$ ↓& $\text{std.}(r_K)$ ↓& $\text{std.}(r_S)$ ↓& $\text{std.}(r_K)$ ↓& $\text{std.}(r_S)$ ↓& $\text{std.}(r_K)$ ↓\\
\midrule
Random & \underline{2.70} & \underline{4.55} & 2.49 & 4.78 & 1.11 & 3.61 & 2.20 & 4.82 \\ 
K-Means & 2.99 & 4.92 & \underline{1.73} & \underline{3.76} & \underline{0.65} & \underline{2.37} & \underline{1.63} & \underline{3.65}\\
\ourmethod{} & \textbf{0.33} & \textbf{1.15} & \textbf{0.66} & \textbf{1.03} & \textbf{0.37}$^\dagger$ & \textbf{0.84}$^\dagger$ & \textbf{0.44} & \textbf{1.66} \\
\bottomrule
\end{tabular}
\caption{Standard deviation(std.) of correlation($r_S, r_K$) in Table~\ref{tab:tinybenchmark}. Compared with random baselines, our method is more stable to select a representative subset.}
\label{tab:std100}
\end{table*}

\begin{table*}[htbp]
\centering
\begin{tabular}{l|cc|cc|cc|cc}
\toprule
\multirow{2}{*}{Method} 
    & \multicolumn{2}{c|}{\textbf{GSM8K} ($k$ = 446)}
    & \multicolumn{2}{c|}{\textbf{ARC} ($k$ = 332)} 
    & \multicolumn{2}{c|}{\textbf{Hellaswag} ($k$ = 125)} 
    & \multicolumn{2}{c}{\textbf{MMLU} ($k$ = 2082)} \\
&  $\text{std.}(r_S)$ ↓ & $\text{std.}(r_K)$ ↓ & $\text{std.}(r_S)$ ↓& $\text{std.}(r_K)$ ↓& $\text{std.}(r_S)$ ↓& $\text{std.}(r_K)$ ↓& $\text{std.}(r_S)$ ↓& $\text{std.}(r_K)$ ↓\\
\midrule
Random & 0.86 & 2.55 & 0.72 & 2.34 & 0.99 & 3.69 & 0.33 & 1.44 \\ 
K-Means & 0.41 & 1.27 & 0.61 & 1.91 & 0.82 & 3.72 & 0.36 & 1.76 \\
Llama3.1 & 0.25 & 1.16 & 0.27 & 1.24 & 0.50 & \underline{1.46} & \underline{0.18} & 1.10\\
Qwen2.5 & \underline{0.20} & \underline{0.68} & \underline{0.18} & \textbf{0.62} & \textbf{0.37} & \textbf{1.44} & 0.45 & \textbf{0.72} \\
Qwen1.5 & \textbf{0.19} & \textbf{0.57} & \textbf{0.13} & \underline{0.91} & \underline{0.42} & 1.57 & \textbf{0.11} & \underline{0.99}\\
\bottomrule
\end{tabular}
\caption{Standard deviation(std.) of correlation($r_S, r_K$) in Table~\ref{tab:dynamic_k} and Table~\ref{tab:ablation_model}. Compared with random baselines, our method is more stable to select a representative subset.}
\label{tab:std_dynamic}
\end{table*}

\begin{table*}[htbp]
\centering
\begin{tabular}{l|cc|cc|cc|cc}
\toprule
\multirow{2}{*}{Method} 
    & \multicolumn{2}{c|}{\textbf{GSM8K} ($k$ = 140)}
    & \multicolumn{2}{c|}{\textbf{ARC} ($k$ = 409)} 
    & \multicolumn{2}{c|}{\textbf{Hellaswag} ($k$ = 149)} 
    & \multicolumn{2}{c}{\textbf{MMLU} ($k$ = 669)} \\
&  $\text{std.}(r_S)$ ↓ & $\text{std.}(r_K)$ ↓ & $\text{std.}(r_S)$ ↓& $\text{std.}(r_K)$ ↓& $\text{std.}(r_S)$ ↓& $\text{std.}(r_K)$ ↓& $\text{std.}(r_S)$ ↓& $\text{std.}(r_K)$ ↓\\
\midrule
Random & 1.37 & \underline{2.87} & \underline{0.45} & \underline{2.19} & \underline{0.51} & 2.06 & 0.91 & \underline{2.66} \\
K-Means & \underline{1.07} & 3.01 & 0.63 & 2.32 & 0.52 & \underline{1.53} & \underline{0.72} & 2.83\\
\ourmethod{} & \textbf{0.26} & \textbf{1.34} & \textbf{0.16} & \textbf{0.57} & \textbf{0.18}$^\dagger$ & \textbf{0.90}$^\dagger$ & \textbf{0.11} & \textbf{0.64} \\
\bottomrule
\end{tabular}
\caption{Standard deviation(std.) of correlation($r_S, r_K$) in Table~\ref{tab:ablation_k}. Compared with random baselines, our method is more stable to select a representative subset.}
\label{tab:std_window}
\end{table*}

\begin{table*}[htbp]
\centering
\begin{tabular}{lcccc}
\toprule
Method
    & \textbf{GSM8K}
    & \textbf{ARC}
    & \textbf{Hellaswag} 
    & \textbf{MMLU} \\
\midrule
Random & 30.1 & 31.6 & 60.8$^\dagger$ & 11.5\\
K-Means & 30.3 & 29.9 & 60.5$^\dagger$ & 10.9 \\
tinyBenchmarks & 30.6 & 33.8 & 61.0$^\dagger$ & 11.0 \\
metabench & 46.0 & 39.5 & 60.7$^\dagger$ & 10.6 \\
\bottomrule
\end{tabular}
\caption{Neuron coverage ratio $r$ of the baselines in Table~\ref{tab:tinybenchmark} and Table~\ref{tab:metabench}. Except for metabench, whose $k$ is set to its originally selected value, all other entries use a fixed $k$ of 100.}
\label{tab:neuron_coverage}
\end{table*}

\begin{table*}[htbp]
\centering
\begin{tabular}{cc}
\toprule
\textbf{Model Series} & \textbf{Model Name} \\
\midrule
\multirow{7}{*}{Qwen} 
  & Qwen-1.5-7B-Chat~\cite{qwen} \\
  & Qwen-2.5-1.5B-Instruct \\
  & Qwen-2.5-7B-Instruct \\
  & Qwen-2.5-14B-Instruct \\
  & Qwen-2.5-32B-Instruct~\cite{qwen2.5} \\
  & Qwen-2.5-Math-7B~\cite{yang2024qwen25mathtechnicalreportmathematical} \\
  & Qwen-3-8B (thinking mode on/off)~\cite{qwen3technicalreport} \\
  & DeepSeek-R1-Distill-Qwen-7B \\
\midrule
\multirow{4}{*}{LLaMA} 
  & Vicuna-7B-v1.3~\cite{vicuna2023} \\
  & LLaMA-2-7B-Chat \\
  & LLaMA-2-13B-Chat~\cite{touvron2023llama} \\
  & LLaMA-3.1-8B-Instruct~\cite{grattafiori2024llama} \\
  & DeepSeek-R1-Distill-LLaMA-8B~\cite{guo2025deepseek} \\
\midrule
\multirow{1}{*}{Gemma} 
  & Gemma-2-9B-it~\cite{gemma_2024} \\
\midrule
\multirow{2}{*}{Proprietary} 
  & GPT-4o-2024-11-20~\cite{openai2024gpt4o} \\
  & Gemini-2.5-Flash-Preview-04-17~\cite{google2025gemini25} \\
\bottomrule
\end{tabular}
\caption{List of evaluated models.}
\label{tab:models}
\end{table*}

\begin{table*}[htbp]
\centering
\begin{tabular}{lccccc}
\toprule
Statistic & GSM8K & ARC & Hellaswag & MMLU & Total \\
\midrule
\# samples & 1319 & 1172 & 10042 & 14042 & 26575\\
\bottomrule
\end{tabular}
\caption{Statistics of datasets.}
\label{tab:dataset}
\end{table*}

\begin{table*}[htb]
\small
\centering
\begin{tabular}{l|ccc|ccc|ccc|ccc}
\toprule
\multirow{2}{*}{Method} 
    & \multicolumn{3}{c|}{\textbf{GSM8K} ($k$ = 446)}
    & \multicolumn{3}{c|}{\textbf{ARC} ($k$ = 332)} 
    & \multicolumn{3}{c|}{\textbf{Hellaswag} ($k$ = 125)} 
    & \multicolumn{3}{c}{\textbf{MMLU} ($k$ = 2082)} \\
&  $r_S$ & $r_K$ & MAE ↓ & $r_S$ & $r_K$ & MAE ↓ & $r_S$ & $r_K$ & MAE ↓ & $r_S$ & $r_K$ & MAE ↓ \\
\midrule
Random & 98.9 & 94.6 & 1.03 & 98.6 & 93.8 & 1.47 & 97.9 & 91.1 & 3.18 & 99.4 & 96.8 & 0.94 \\
K-Means & 99.2 & 96.1 & 1.40 & 98.7 & 94.3 & 1.52 & 98.2 & 92.0 & 3.11 & 99.7 & 98.0 & 1.97 \\
\midrule
Llama3.1-8B & 99.3 & 96.3 & 1.99 & 99.0 & 95.6 & 1.48 & 95.1$^-$ & 86.7$^-$ & 2.94$^-$ & 99.8 & 98.5 & 0.94 \\ 
Qwen2.5-7B & 98.9 & 94.7 & 2.32 & 98.8 & 94.5 & 2.76 & 98.5 & 92.3 & 2.78 & 96.9 & 88.8 & 10.85\\
Qwen1.5-7B & 99.1 & 95.2 & 1.09 & 98.9 & 94.9 & 1.90 & 99.0 & 95.9 & 4.23 & 97.9 & 91.2 & 9.41 \\
\bottomrule
\end{tabular}
\caption{Comparison between different indicator models on all four benchmarks. Due to safety constraints, LLaMA refuses to answer certain samples in Hellaswag (marked with $^-$), making it unable to effectively guide the
data selection process.}
\label{tab:ablation_model_full}
\end{table*}

\end{document}